\documentclass[10pt,twocolumn,letterpaper]{article}

\usepackage[pagenumbers]{cvpr} 

\usepackage{graphicx}
\usepackage{amsmath}
\usepackage{amssymb}
\usepackage{booktabs}
\usepackage{makecell}

\usepackage{pifont}
\usepackage{lipsum}
\usepackage{hhline}
\usepackage{multirow}
\usepackage{tabularx}
\usepackage{color}
\usepackage[cmyk]{xcolor}
\usepackage[pagebackref,breaklinks,colorlinks]{hyperref}

\usepackage[capitalize]{cleveref}
\crefname{section}{Sec.}{Secs.}
\Crefname{section}{Section}{Sections}
\Crefname{table}{Table}{Tables}
\crefname{table}{Tab.}{Tabs.}

\def\confName{CVPR}
\def\confYear{2024}
\def\paperID{2837}

\begin{document}

\title{MPTQ-ViT: Mixed-Precision Post-Training Quantization for Vision Transformer}

\author{Yu-Shan Tai, An-Yeu (Andy) Wu\\
Graduate Institute of Electrical Engineering, National Taiwan University\\
Taipei, Taiwan \\
{\tt\small clover@access.ee.ntu.edu.tw, andywu@ntu.edu.tw}
\and
\\
}
\maketitle

\begin{abstract}
    While vision transformers (ViTs) have shown great potential in computer vision tasks, their intense computation and memory requirements pose challenges for practical applications.
    Existing post-training quantization methods leverage value redistribution or specialized quantizers to address the non-normal distribution in ViTs. 
    However, without considering the asymmetry in activations and relying on hand-crafted settings, these methods often struggle to maintain performance under low-bit quantization.  
    To overcome these challenges, we introduce SmoothQuant with bias term (SQ-b) to alleviate the asymmetry issue and reduce the clamping loss.
    We also introduce optimal scaling factor ratio search (OPT-m) to determine quantization parameters by a data-dependent mechanism automatically.
    To further enhance the compressibility, we incorporate the above-mentioned techniques and propose a mixed-precision post-training quantization framework for vision transformers (MPTQ-ViT).
    We develop greedy mixed-precision quantization (Greedy MP) to allocate layer-wise bit-width considering both model performance and compressibility.
    While previous works focus on individual challenge, our work is the first to simultaneously address issues of asymmetry, non-normal distribution, and uneven sensitivity.
    Our experiments on ViT, DeiT, and Swin demonstrate significant accuracy improvements compared with SOTA on the ImageNet dataset.
    Specifically, our proposed methods achieve accuracy improvements ranging from 0.90\% to 23.35\% on 4-bit ViTs with single-precision and from 3.82\% to 78.14\% on 5-bit fully quantized ViTs with mixed-precision.

\end{abstract}

\section{Introduction}\label{sec:intro}
    \begin{figure*}[t]
        \centering
        \includegraphics[width=2\columnwidth]{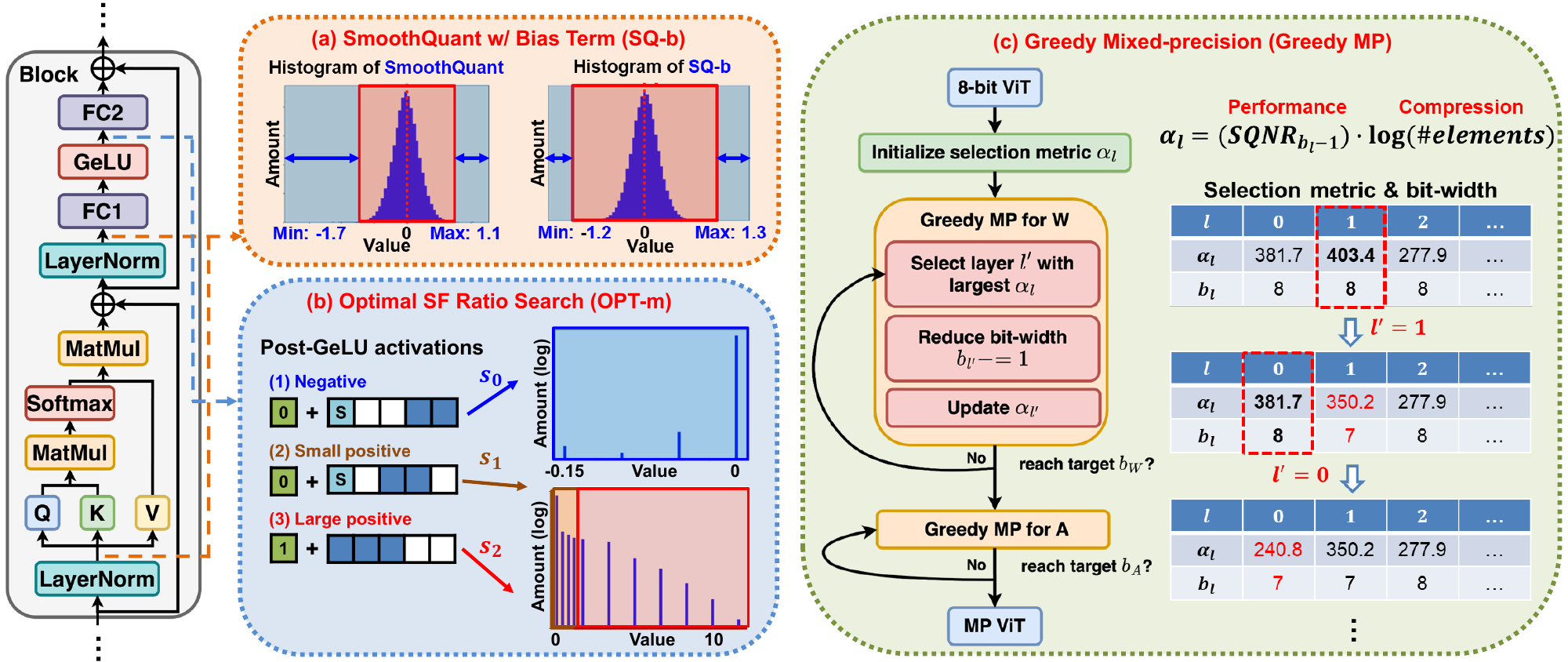}
        \vspace{-0.7em}
        \caption{Proposed mixed-precision post-training quantization framework for ViT (MPTQ-ViT). (a) SQ-b, (b) OPT-m, and (c) Greedy MP.}
       \label{fig:Overall}
       \vspace{-1.3em}
    \end{figure*}
    Initially intended for natural language processing (NLP) tasks \cite{bert}, transformer-based models have expanded their applications to various computer vision (CV) tasks and achieved remarkable performance \cite{vit, deit, Swin, mask, cascade, segformer}. 
    However, vision transformers (ViT) usually demand substantial parameters and computational consumption. 
    For instance, ViT-L \cite{vit} consists of 307M parameters and requires 64G FLOPs.  
    The excessive overheads hinder ViTs from embedding on resource-constrained edge devices. 
    Consequently, model compression for ViTs has received considerable attention recently.
    Common model compression techniques include quantization \cite{up_down, opt_exponent, alps,learnablMP,rethinking,GMPQ,evolutionary, ptq_for_vit,data_free_MP}, pruning \cite{apq,deep_com,xpruner,joint_pruning}, and dimension reduction \cite{GreedyDR, feature_trans, joint_mp_dr}. Among these approaches, quantization is widely used due to its simplicity and practicality.
    
    By converting floating-point (FP) values to discrete integers, quantization reduces memory requirements and computational overheads. 
    There are two types of quantization methods, quantization-aware training (QAT) \cite{qvit,binaryvit,tervit,Learnable,bibert,qbert} and post-training quantization (PTQ) \cite{ptq_for_vit, ptq4vit, pd_quant, tsptq_vit, smoothquant, outlier_suppression+, apq_vit, fqvit}. 
    Incorporating quantization loss during training, QAT enables aggressive low bit-width (BW) models. 
    However, the requirements of retraining with labeled datasets confine its applications.
    Alternatively, PTQ employs a small amount of unlabeled data to calibrate the quantization parameters, significantly accelerating the process. Existing PTQ methods are primarily designed for Convolutional Neural Networks (CNNs) \cite{up_down, opt_exponent, alps, learnablMP, rethinking,evolutionary}. 
    Without considering the unique characteristics of ViTs, applying these methods directly leads to unacceptable accuracy. 
    Consequently, some researchers have modified the non-normal distribution in transformers to favor quantization. 
    The authors of SmoothQuant \cite{smoothquant} introduce a smoothing factor to transfer the quantization difficulty from activations to weights. 
    On the other hand, some other researchers have developed specialized quantizers to mitigate the quantization loss. 
    The authors of PTQ4ViT \cite{ptq4vit} introduce a region-specific quantization scheme to address the non-normal distribution of post-Softmax and post-GeLU values. 
    TSPTQ-ViT \cite{tsptq_vit} further considers the high channel-wise variance of LayerNorm and proposes value-aware and outlier-aware two-scaled scaling factors.

    Different from the above-mentioned methods using single-precision quantization (SP) to assign the same BW across the entire network, mixed-precision quantization (MP) utilizes fine-grained BW allocation to further exploit model redundancy \cite{learnablMP,rethinking,GMPQ,ptq_for_vit,evolutionary,data_free_MP}. 
    However, research on post-training MP for ViTs is limited. 
    To the best of our knowledge, only Liu \etal \cite{ptq_for_vit} have used the Pareto frontier approach with a nuclear norm-based metric to achieve module-wise MP.
    
    Though the aforementioned methods offer solutions to certain challenges of PTQ for ViTs, there are still several issues that need to be addressed:

    \vspace{-0.2em}
    \begin{enumerate}
        \item \textit{Asymmetric distribution among activations}: 
        While SmoothQuant \cite{smoothquant} mitigates the outlier issues, the asymmetry in activations remains a challenge for symmetric quantization. This results in sub-optimal data representation and potential accuracy degradation.
        \item \textit{Manually-designed quantizer for post-GeLU values}: 
        Existing approaches \cite{ptq4vit, tsptq_vit} rely on manually-designed rules to refine the quantizer. 
        Though these methods perform effectively under 8-bit quantization, they encounter challenges when extending to low-bit scenarios.
        \item \textit{Neglecting diverse layer-wise sensitivity}: 
        Most PTQ methods for ViTs focus solely on single-precision designs \cite{ptq4vit, pd_quant, tsptq_vit, smoothquant, outlier_suppression+, apq_vit, fqvit}, overlooking the sensitivity to quantization varies with different layers and operator types.
    \end{enumerate}
    
    \vspace{-0.2em}
    To address the above issues, we propose a novel mixed-precision post-training quantization framework for ViTs (MPTQ-ViT), as shown in \cref{fig:Overall}. 
    Our main contributions are as follows:
    
    \vspace{-0.2em}
    \begin{enumerate}
        \item \textit{SmoothQuant with Bias Term (SQ-b)}: 
        Building upon SmoothQuant \cite{smoothquant}, we introduce a bias term to alleviate the asymmetry in activations. 
        The entire process can seamlessly integrate into the original network without additional inference overheads.
        \item \textit{Optimal Scaling Factor Ratio Search (OPT-m)}: 
        We propose automatically deciding the scaling factors (SFs) for post-GeLU values by a data-dependent mechanism, eliminating manual intervention and improving adaptability.
        \item \textit{Greedy Mixed-Precision Quantization (Greedy MP)}: 
        We design a selection metric considering both model performance and compressibility. 
        Guided by this metric, we adopt a greedy strategy to efficiently determine the layer-wise BW of weights and activations.
    
    \end{enumerate}                     

\section{Related Works}\label{sec:related works}
    \subsection{Vision Transformer (ViTs)}
        Transformer-based models have excelled in NLP tasks and successfully transferred to CV fields, e.g., image classification \cite{vit, deit,Swin}, object detection \cite{mask,cascade}, and segmentation \cite{segformer}.
        While CNNs rely solely on local pixel information, ViTs leverage self-attention mechanisms \cite{attention} to extract global information. 
        ViT \cite{vit} is the first work to send image patches into the transformer-based models, resulting in impressive performance across multiple image classification tasks. 
        However, the considerable number of parameters and quadratic computational costs limit practical applications. 
        Consequently, recent research has focused on designing lightweight and efficient variants. 
        For instance, DeiT \cite{deit} utilizes a distillation-based approach to achieve competitive performance with limited data, while Swin \cite{Swin} introduces a hierarchical architecture and shifted attention mechanisms to incorporate local and global information and enhance efficiency. 
        Unlike these model architecture refinements, our work focuses on quantization techniques applicable to various ViTs.
        
    \subsection{Value Redistribution for Quantization}
        The non-normal distribution in transformers poses a challenge for quantization. 
        Consequently, some researchers have explored value redistribution before quantization.
        Observing the systematic outliers in fixed activation channels, the authors of SmoothQuant \cite{smoothquant} utilize a channel-wise smoothing factor $\epsilon$ to redistribute the weights and activations.
        To maintain the mathematical equivalence without extra scaling, the smoothing factors are off-line fused into linear operators, \ie LayerNorm and linear layers.
        Specifically, the output of LayerNorm can be specified as the linear transpose of the normalized inputs $Y=\Bar{X} \cdot \gamma + \beta$. 
        Then, suppose the following linear layer, QKV or FC1, is weighted by $W$ and $b$. The equivalent transformation can be specified as:
        \vspace{-0.5em}
        \begin{equation}
            Y\cdot W^{T}+b=\frac{Y}{\epsilon}\cdot \epsilon W^{T}+b,
            \label{eq:def_SQ}
        \vspace{-0.3em}
        \end{equation}
        where the smoothing factor is channel-wisely set $\epsilon_j=({max(Y_j)}/{max(W_j)})^{0.5}$ for each channel $j$.
        Though alleviating the outlier issues, the asymmetry in activations is still challenging to symmetric quantization. 
        The histogram after SmoothQuant \cite{smoothquant} in \cref{fig:Overall}(a), obtained by the inputs of the last FC1 in DeiT-S,  reveals that the negative range is wider than the positive range.
        Symmetric quantization allocates an equal range for both positive and negative values, as depicted in the red region.
        This imbalance leads to the clamping of many large-magnitude values in the blue region and thus brings huge losses.
        To alleviate this issue, Outlier Suppression+ \cite{outlier_suppression+} introduces a shifting operation to eliminate the asymmetry.
        NoisyQuant \cite{noisyquant} is the first to transfer the techniques from NLP to ViTs, deploying additive noisy bias to heavy-tailed activation distribution. 
        However, \cite{noisyquant} incurs additional inference overhead due to the summation of inputs and additive noise.
        Thus, we aim to address the asymmetry in ViTs without additional computation.
        
    \subsection{Specialized Post-Training Quantizer for ViTs}
        Parallel to value redistribution, some researchers have focused on specialized quantizers to address certain non-normal distributed activations in ViTs.
        While QAT \cite{qvit,binaryvit,tervit,Learnable,bibert,qbert} involves extensive fine-tuning, PTQ \cite{ptq_for_vit, ptq4vit, pd_quant, tsptq_vit, smoothquant, outlier_suppression+, apq_vit, fqvit} relies on a small calibration dataset to determine quantization parameters. 
        Considering hardware support for quantization, uniform symmetric quantization is the most commonly adopted quantizer. 
        For a floating-point value \textit{x}, The \textit{b}-bit representation $x_{q,b}$ and the reconstruction value $\hat{x}_{q,b}$ can be obtained through a uniform scaling factor (SF) \textit{s}:
        \begin{equation}
            x_{q,b} = clamp(\lfloor x / s\rceil , -2^{b-1},2^{b-1}-1),
            \label{eq:def_quant}
        \end{equation}
        \begin{equation}
             \hat{x}_{q,b} = s \cdot x_{q,b},
            \label{eq:def_quant_rec}
        \end{equation}
        where \textit{clamp(x,l,h)} is used to clamp \textit{x} in [\textit{l,h}] and $\lfloor\cdot\rceil$ is the round-to-nearest operator. 
        However, the naive approach is unsuitable for the non-normal distributed activations in ViTs. 
        For instance, the post-GeLU activations exhibit unbalanced positive and negative ranges, as shown in Neg-GeLU and Pos-GeLU of \cref{fig:OPTm}.
        Consequently, some researchers have proposed optimized quantizers.  
        Liu \etal \cite{ptq_for_vit} were pioneers in employing PTQ to ViTs, leveraging a ranking loss and similarity metric to search optimal SFs. 
        FQ-ViT \cite{fqvit} first achieves 8-bit fully quantized ViT by power-of-two factors and log2 quantization for LayerNorm and Softmax operations. 
        The authors of APQ-ViT \cite{apq_vit} utilize block-wise calibration and preserve the Matthew effect of post-Softmax values. 
        In PTQ4ViT \cite{ptq4vit}, the authors propose the Hessian guided metric to evaluate the quantization impact and region-specific quantizer for activations after Softmax and GeLU. 
        Moreover, TSPTQ-ViT \cite{tsptq_vit} proposes two-scaled quantizers, V-2SF for post-Softmax and GeLU values, and O-2SF for LayerNorm. 
        Despite achieving near lossless performance under 8-bit quantization, the hand-crafted designs in \cite{ptq4vit,tsptq_vit} encounter challenges when dealing with the non-normal distributed post-GeLU values under low-bit quantization.
        To alleviate this issue, we propose automatically determining suitable SFs in a data-dependent manner.

    \subsection{Mixed-Precision Quantization (MP)}
        While single-precision quantization (SP) assigns a uniform BW to the entire network, MP  \cite{learnablMP,rethinking,GMPQ,ptq_for_vit,evolutionary,data_free_MP} leverages fine-grained BW allocation to enhance redundancy utilization. 
        However, determining the optimal BW allocation is challenging due to the vast and non-differentiable search space. 
        Learning-based MP methods \cite{learnablMP, rethinking,GMPQ}, with differentiable search, often incur substantial time and memory overhead. 
        In contrast, post-training MP \cite{evolutionary,ptq_for_vit,data_free_MP} approaches rely on well-designed metrics to guide the quantization process. 
        Nevertheless, the research on post-training MP for ViTs is currently limited.
        To the best of our knowledge, only Liu \etal \cite{ptq_for_vit} use the nuclear norm and Pareto frontier to achieve module-wise MP, assigning all the operators in the multi-head self attention (MSA) or multi layer perception (MLP) modules with the same BW.
        Since both operator types and locations impact quantization sensitivity, module-wise MP is insufficient.
        Instead, we propose a layer-wise MP strategy to effectively address the various quantization requirements.

\section{Proposed Methods}\label{sec:methods}

    \subsection{SmoothQuant with Bias Term (SQ-b)} \label{sec: SQ-b}
        The non-normal distributed activations in ViTs induces outlier and asymmetry issues, thus causing inefficient representation and intolerable clamping loss.
        To better preserve the model precision, we propose to redistribute values to favor symmetric quantization.
        To avoid extra inference overheads, we follow \cite{smoothquant} to operate equivalent transformation on the post-LayerNorm values.  
        Different from \cref{eq:def_SQ}, we introduce a bias term $\mu$ to mitigate the asymmetry issue:
        \vspace{-0.5em}
        \begin{equation}
            Y\cdot W^{T}+b=\frac{(Y - \mu)}{\epsilon}\cdot \epsilon W^{T}+b +\mu W^{T}
            \label{eq:def_SQb1}
        \end{equation}
        \vspace{-0.5em}
        \begin{equation}
            =(\Bar{X} \cdot \gamma_{new} + \beta_{new}) \cdot W_{new}^{T}+b_{new}. 
            \label{eq:def_SQb2}
        \end{equation}
        Replacing the original weights with adjusted values, the transformation maintains FP equivalence: $\gamma_{\text{new}}=\frac{\gamma}{\epsilon}, \beta_{\text{new}}=\frac{\beta-\mu}{\epsilon}, W_{\text{new}}^{T}=\epsilon {W}^{T}, b_{\text{new}}=b+\mu W^{T}$.
        For each channel $j$, we set $\mu_j$ as $mean(Y_j )$ to translate the activations to zero means.
        The histogram in \cref{fig:Overall}(a) illustrates the impact of applying SQ-b, resulting in a noticeable reduction in the clamping range (blue region) and thus mitigating the clamping loss.
        
        With the proposed SQ-b, the redistributed values are more suitable for symmetric quantization, benefiting the subsequent quantization process.
        In \cref{sec:analysis_SQb}, we would further analyze different settings of $\epsilon$ and $\mu$.

    \subsection{Optimal Scaling Factor Ratio Search (OPT-m)} \label{sec: OPT-m}
        \begin{figure}[t]
            \centering
            \includegraphics[width=1\linewidth]{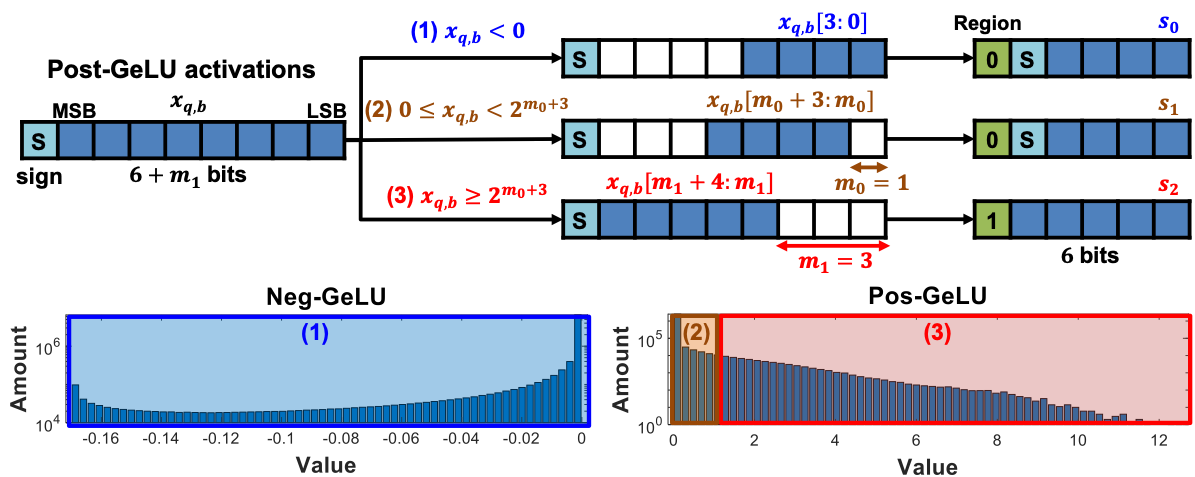}
            \vspace{-2em}
            \caption{OPT-m under 6-bit quantization. Neg-GeLU/Pos-GeLU are the histograms of negative/positive post-GeLU values.}
           \label{fig:OPTm}
            \vspace{-1.8em}
        \end{figure}
        Though SQ-b addresses the asymmetry issues, there are still challenges in quantizing ViTs. 
        The post-GeLU values have an unbounded positive range, while the negative range is concentrated within a narrow range of less than 0.2, as shown in \cref{fig:OPTm}.
        Additional post-GeLU distribution visualizations can be found in the Appendix.
        The non-normal distribution makes the existing hand-crafted quantizers \cite{ptq4vit,tsptq_vit} unable to accurately represent the post-GeLU values under low-bit quantization.
        To address this, we propose Optimal Scaling Factor Ratio Search (OPT-m), a data-dependent strategy to automatically determine the region-specific SFs.
        This approach involves systematically solving two key problems.

        The first problem is \textit{how to split the regions of post-GeLU values}.
        We introduce two data-dependent hyperparameters, $m_0$ and $m_1$, to separate the post-GeLU values into three regions: (1) negative values, (2) small positive values, and (3) large positive values. 
        Each region is assigned a specific SF, \{$s_0, s_1, s_2$\}, respectively. 
        To be hardware-friendly, these SFs can be fast aligned by shifting $m_0$ and $m_1$ bits:
        \vspace{-0.3em}
        \begin{equation}
            s_1 = s_0\cdot2^{m_0}, s_2 = s_0\cdot2^{m_1}.
            \label{eq:def_m}
        \vspace{-0.3em}
        \end{equation}
        We consider the 6-bit quantization shown in \cref{fig:OPTm} as an example.
        The value is initially quantized to $x_{q,b}$ with $b=(6+m_1)$ and $s=s_0$ by \cref{eq:def_quant}, where the first bit denotes the sign and the others represent the magnitude.
        The regions are distinguished by region and sign indices.
        Region (1)(2) share the region index of 0 and are distinguished by a sign bit, while region (3) is recognized by the region index of 1.
        Apart from the region and sign indices, there remain 4 bits for region (1)(2) and 5 bits for region (3) to record the quantized values.
        Then, we record different portions of consecutive quantized bits for each region, with the first truncated bit being rounded.
        For region (1), due to the small magnitude, those values can be presented just by the least significant few bits. 
        Thus, combined with the leading 2 bits denoting the region and sign indices, the presentation is $[0,1,x_{q,b}[3:0]]$ with SF of $s_0$.
        Similarly, for region (2), values satisfying $0\leq x_{q,b}<2^{m_0+3}$ are represented by $[0,0,x_{q,b}[m_0+3:m_0]]$ with SF of $s_1$. 
        As for region (3), those values larger than $2^{m_0+3}$ is formatted as $[1,x_{q,b}[m_1+4:m_1]]$ with SF of $s_2$. 
        After quantization, the quantized values can be restored by multiplying with their region-specific SFs. 
        The three regions are separated by $0$ and $2^{m_0+3}$. Since $m_0$ is a data-dependent parameter, we break the limitation of hand-crafted region division.

        The other problem is \textit{how to determine the suitable SFs}, \ie $m_0$, $m_1$, and $s_0$. 
        To well represent the variant post-GeLU values, we determine $m_0$ and $m_1$ by analyzing the data distribution.
        We explain the process by 6-bit quantization, as shown in \cref{fig:OPTm}. 
        The quantized values reach maximum magnitude when all the reserved quantized bits are set to 1.  
        Thus, the most negative value is $-15\cdot s_0$ in the region (1), while the most positive one is $31\cdot s_2$ in the region (3).
        To establish the lower and upper bounds of the quantizer, $x_{low}$ and $x_{up}$, we use the average minimum value and the 99.95\% percentile of the post-GeLU values obtained from the calibration data.
        Consequently, we can assume the optimal $s_0$ and $s_2$ will be close to $\frac{x_{low}}{-15}$ and $\frac{x_{up}}{31}$, respectively.
        Given the relationship specified in \cref{eq:def_m}, we determine the value of $m_1$ by: 
        \vspace{-0.5em}
        \begin{equation}
            m_1 = \lfloor \log_2 \frac{s_2}{s_0} \rceil 
                = \lfloor \log_2 {\frac{x_{up}}{31}}/{\frac{x_{low}}{-15}} \rceil.
            \label{eq:def_m1}
        \vspace{-0.3em}
        \end{equation}
        After obtaining $m_1$, we restrict the candidates of $m_0$ to be less than $m_1$, as the SF of small positive values should be smaller than that of large positive values. We then determine the desired $m_0$ by minimizing the Hessian-guided metric \cite{ptq4vit} among these candidates:
        \begin{equation}
        \vspace{-0.3em}
            m_0 = \mathop{\arg\min}\limits_{m} \{Hessian(m)|m\in [0,1,...,m_1-1]\}.
            \label{eq:def_m0}
        \vspace{-0.3em}
        \end{equation}
        Finally, we also apply the Hessian-guided metric \cite{ptq4vit} to linear divided candidates to find the base SF, $s_0$, and calculate $s_1$ and $s_2$ according to \cref{eq:def_m}. 
        
        Following the above-mentioned rules, we can obtain the data-dependent SFs, enhancing the adaptability compared to manual methods. In \cref{sec:analysis_OPTm}, we further visualize the post-quantization distribution to evaluate the effectiveness of OPT-m.
        
    \subsection{Greedy MP Quantization (Greedy MP)}
        Existing PTQ methods for ViTs primarily focus on single-precision quantization (SP), overlooking the varying sensitivity to quantization across layers. To further improve compression efficiency, we propose the Greedy Mixed-Precision Quantization (Greedy MP) to allocate specific BW to individual layers.

        During the quantization process, two crucial factors should be considered: \textit{model performance} and \textit{achieved compression}.
        Model performance refers to how well the quantized model retains the functionality compared to the original FP model, while achieved compression refers to the reduction in model size through quantization. 
        Achieving a balance between performance and compressibility is important. 
        Specifically, aggressive quantization can result in significant compression gains but may degrade model performance.
        Conversely, conservative quantization may preserve model performance but limit compression benefits.
        
        Thus, we need a metric considering both factors. 
        Since obtaining the exact model accuracy is time-consuming and often unavailable,  an alternative metric is necessary. 
        As stated in \cite{alps, opt_exponent}, a linear relationship exists between model accuracy and signal-to-quantization-noise ratio (SQNR). 
        The SQNR of a matrix $X$ under \textit{b}-bit quantization can be efficiently calculated by dividing the power spectrum of the signal by the quantization noise:
        \vspace{-0.8em}
        \begin{equation}
            SQNR_b(X) = 10\cdot log(\frac{\sum X^2}{\sum (X-\hat{X}_{q,b})^2}).
            \label{eq:def_sqnr}
        \vspace{-0.4em}
        \end{equation}
        As for the achieved compression, we quantify it using the logarithm of the number of elements.
        The logarithmic term prevents the range of compression benefits from being overly disproportionate to SQNR.
        Finally, we calculate the selection metric $\alpha_l$ of layer $l$ to assess its suitability for further quantization to $b$-1 bits by:
        \begin{equation}
        \vspace{-0.3em}
            \alpha_l = SQNR_{b-1}(X_l)\cdot log (numel(X_l)).
            \label{eq:def_metric}
        \vspace{-0.2em}
        \end{equation}
        $X_l$ represents either weights or activations, depending on the quantization target, and $numel(X_l)$ denotes the number of elements in $X_l$.
        Larger $\alpha_l$ indicates the potential for higher performance and greater compression benefits.
        
        With the selection metric to evaluate quantization priority, we then apply a greedy strategy to iteratively reduce the bit-width in a layer-wise manner. 
        We aim to obtain the MP model satisfying the target bit-widths of weights and activations, $b_W$ and $b_A$.
        The flow of Greedy MP is shown in \cref{fig:Overall}(c), which contains three steps:
        \begin{enumerate}
            \item \textit{Initialization}: 
            We start with ViTs under W8A8 SP and initialize the selection metric of each layer by \cref{eq:def_metric} to assess the potential impact of further quantization.
            \item \textit{Greedy MP for weights}: 
            We identify the target layer $l^{'}$ with the largest $\alpha_l$. Then, we substrate the BW $b_{l^{'}}$ by 1 and update $\alpha_{l^{'}}$ accordingly. This step is repeated until the average BW of weights reaches the target BW $b_W$.  
            \item \textit{Greedy MP for activations}: After quantizing the weights, we perform the same process on activations until reaching the target bit-width $b_A$. Note that exchanging steps 2 and 3 results in similar performances and thus can be substituted.
        \end{enumerate}
        
       Finally, after the iterative BW reduction, the Greedy MP enables layer-wise bit-width allocation considering both model performance and achieved compression.
        In \cref{sec:analysis_MP}, we further analyze the results obtained from Greedy MP to identify the key factors influencing the outcomes.

        The three proposed techniques, \ie SQ-b, OPT-m, and Greedy MP, can be operated independently or jointly. To validate their combined benefits, we integrate them into the MPTQ-ViT framework. 
        In this setup, the FP model initially employs SQ-b to adjust the value distribution and subsequently undergoes the Greedy MP process.
        During the \textit{Greedy MP for activations} step, we employ OPT-m if post-GeLU activations are selected as layer $l^{'}$.
        \begin{table*}[t]
            \centering
            \begin{tabular}{@{}c|ccccccccccc@{}}
            \Xhline{1pt}
            Method      & W/A   & \#img & ViT-S & ViT-B & ViT-L & DeiT-T & DeiT-S & DeiT-B & Swin-T & Swin-S & Swin-B \\ \hhline{============}
            FP          & 32/32 & N/A     & 81.39 & 84.53 & 85.84 & 72.18  & 79.85  & 81.80  & 81.37  & 83.21  & 85.27  \\ \hline
            Liu.\cite{ptq_for_vit}        & 6/6   & 1024  & -     & 75.26 & 75.46 & -      & 74.58  & 77.02  & -      & -      & -      \\
            PTQ4ViT\cite{ptq4vit}     & 6/6   & 32    & 78.63 & 81.65 & 84.79 & 69.62  & 76.28  & 80.25  & 80.47  & 82.38  & 84.01  \\
            PD-Quant\cite{pd_quant}    & 6/6   & 1024  & 70.84 & 75.82 & -     & -      & 78.33  & -      & -      & -      & -      \\
            APQ-ViT\cite{apq_vit}     & 6/6   & 32    & 79.10 & 82.21 & -     & 70.49  & 77.76  & 80.42  & -      & 82.67  & 84.18  \\
            NoisyQuant\cite{noisyquant} & 6/6   & 1024  & 78.65 & 82.32 & -     & -      & 77.43  & 80.70  & 80.51  & \textbf{82.86}  & \textbf{84.68}  \\
            TSPTQ-ViT\cite{tsptq_vit}   & 6/6   & 32    & 79.45 & 82.29 & 85.18 & 70.82  & 77.18  & 80.61  & 80.62  & 82.60  & 84.16  \\
            \textbf{SQ-b+OPT-m} &
              6/6 &
              32 &
              \textbf{79.98} &
              \textbf{82.70} &
              \textbf{85.53} &
              \textbf{71.03} &
              \textbf{78.70} &
              \textbf{81.25} &
              \textbf{80.67} &
              82.62 &
              84.50 \\ \hline
            PTQ4ViT\cite{ptq4vit}     & 4/4   & 32    & 42.57 & 30.69 & 78.38 & 36.96  & 34.08  & 64.39  & 73.48  & 76.09  & 74.02  \\
            APQ-ViT\cite{apq_vit}     & 4/4   & 32    & 47.95 & 41.41 & -     & 47.94  & 43.55  & 67.48  & -      & 77.15  & 76.48  \\
            TSPTQ-ViT\cite{tsptq_vit}   & 4/4   & 32    & 52.56 & 50.10 & 77.64 & 48.36  & 45.08  & 69.45  & 72.48  & 76.30  & 73.28  \\
            \textbf{SQ-b+OPT-m} &
              4/4 &
              32 &
              \textbf{55.88} &
              \textbf{61.84} &
              \textbf{80.07} &
              \textbf{55.62} &
              \textbf{68.43} &
              \textbf{76.14} &
              \textbf{73.82} &
              \textbf{77.20} &
              \textbf{76.51} \\ \Xhline{1pt}
            \end{tabular}
            \vspace{-0.8em}
            \caption{Top-1 accuracy under single-precision quantization (SP) on ImageNet dataset. Softmax and LayerNorm are preserved in FP. W/A denotes the bit-width for weight/activation, and \#img represents the size of calibration data.}
            \label{tab:SP}
            \vspace{-1.7em}
        \end{table*}
        
        \begin{table}[t]
        \centering
            \begin{tabular}{c|cccc}
            \Xhline{1pt}
            Method          & W/A   & ViT-B          & DeiT-S         & Swin-B         \\ \hhline{=====}
            FP              & 32/32 & 84.53          & 79.85          & 85.27          \\ \hline
            Baseline        & 4/4   & 50.10          & 45.08          & 73.28          \\
            SQ \cite{smoothquant}    & 4/4   & 52.31          & 64.22          & 76.31          \\
            SQ-b  & 4/4   & 55.28          & 67.33          & 75.44          \\
            \textbf{SQ-b+OPT-m} & 4/4   & \textbf{61.84} & \textbf{68.43} & \textbf{76.51}\\ \Xhline{1pt}
            \end{tabular}
            \vspace{-0.8em}
            \caption{Ablation study under SP on ImageNet dataset.}
            \label{tab:ablation}
            \vspace{-1.9em}
        \end{table}
\section{Experimental Results}\label{sec:exp}
    In the following experiments, we evaluate the effectiveness of the proposed methods on different tasks and provide in-depth analyses.  
    For image classification, we conduct experiments on the ImageNet (ILSVRC 2012) \cite{imagenet} dataset using pre-trained ViT \cite{vit}, DeiT \cite{deit}, and Swin \cite{Swin} from timm \cite{timm}.
    As for object detection, we implement on COCO dataset \cite{coco} using Mask R-CNN \cite{mask} and Cascade Mask R-CNN \cite{cascade} with Swin \cite{Swin} as the backbone.
    We choose TSPTQ-ViT \cite{tsptq_vit} as our baseline.
    To maintain model performance under low BW quantization, we modify the input parameter $k$ in the K-means algorithm of O-2SF \cite{tsptq_vit} from 2 to 4.
    Except for this modification, we follow the same simulation settings as in \cite{tsptq_vit}.
    
    For calibration, we randomly select 32 images from the ImageNet dataset for classification and only 1 image from the COCO dataset for object detection.
    The candidates of $s_0$ are 100 equally spaced values by linearly dividing [0, 1.2$s$], where $s$ is set as $max(|X|)/{2^{b-1}}$.
    We apply quantization to all the weights and activations involved in fully connected layers and matrix multiplication.
    While prior works \cite{ptq_for_vit, ptq4vit,pd_quant,apq_vit,noisyquant} just optimize the post-Softmax and post-Layernorm activations but still calculate the two operators in floating-point, our system supports fully quantized ViTs.
    In \cref{sec:performance}, we evaluate our methods in both SP and MP scenarios.
    To ensure a fair comparison, Softmax and LayerNorm are preserved in FP when compared with prior methods.
    For the validation of fully quantized ViTs, we consider the Baseline method mentioned above as the competitor.
    In \cref{sec:analysis_SQb,sec:analysis_OPTm,sec:analysis_MP}, we provide in-depth analyses of the proposed techniques.

        \linespread{0.95}
        \begin{table*}[t]
            \centering
            \begin{tabular}{c|ccccccccccc}
            \Xhline{1pt}
            Method &
              W/A &
              \#img &
              ViT-S &
              ViT-B &
              ViT-L &
              DeiT-T &
              DeiT-S &
              DeiT-B &
              Swin-T &
              Swin-S &
              Swin-B \\\hhline{============}
            FP &
              32/32 &
              N/A &
              81.39 &
              84.53 &
              85.84 &
              72.18 &
              79.85 &
              81.80 &
              81.37 &
              83.21 &
              85.27 \\ \hline
            Baseline (SP) &
              6/6&
              32 &
              78.68 &
              79.57 &
              84.57 &
              70.34 &
              76.78 &
              78.95 &
              45.54 &
              41.95 &
              37.82 \\
            {Greedy MP} &
              6/6&
              32 &
              79.51 &
              82.90 &
              85.43 &
              70.79 &
              78.83 &
              81.06 &
              \textbf{73.50} &
              62.86 &
              60.02 \\
            \textbf{MPTQ-ViT} &
              6/6 &
              32 &
              \textbf{79.90} &
              \textbf{83.12} &
              \textbf{85.58} &
              \textbf{71.10} &
              \textbf{79.29} &
              \textbf{81.29} &
              73.08 &
              \textbf{63.08} &
              \textbf{60.18} \\ \hline
            Baseline (SP) &
              5/5&
              32 &
              3.60 &
              0.30 &
              80.02 &
              26.08 &
              54.74 &
              58.53 &
              33.83 &
              31.00 &
              8.74 \\
            {Greedy MP} &
              5/5 &
              32 &
              55.95 &
              \textbf{79.66} &
              83.61 &
              56.79 &
              73.91 &
              76.44 &
              55.58 &
              39.02 &
              41.69 \\
            \textbf{MPTQ-ViT} &
              5/5 &
              32 &
              \textbf{74.65} &
              78.44 &
              \textbf{83.84} &
              \textbf{57.48} &
              \textbf{76.56} &
              \textbf{77.48} &
              \textbf{56.17} &
              \textbf{42.86} &
              \textbf{44.17}\\ \Xhline{1pt}
            \end{tabular}
            \vspace{-0.8em}
            \caption{Top-1 accuracy under mixed-precision quantization (MP) on ImageNet dataset. MPTQ-ViT includes Greedy MP+SQ-b+OPT-m. All modules are quantized including Softmax and LayerNorm. W/A denotes the bit-width for weight/activation.}
            \label{tab:MP}
            \vspace{-0.8em}
        \end{table*}
        \linespread{1.0}

        \linespread{0.95}
        \begin{table*}[t]
        \centering
        \begin{tabular}{c|ccccccccc}
        \Xhline{1pt}
        \multirow{3}{*}{Method} &
          \multirow{3}{*}{W/A} &
          \multicolumn{2}{c}{Mask R-CNN} &
          \multicolumn{2}{c}{Mask R-CNN} &
          \multicolumn{2}{c}{Cascade Mask R-CNN} & 
          \multicolumn{2}{c}{Cascade Mask R-CNN} \\ 
         &
           &
          \multicolumn{2}{c}{Swin-T} &
          \multicolumn{2}{c}{Swin-S} &
          \multicolumn{2}{c}{Swin-T} &
          \multicolumn{2}{c}{Swin-S} \\
         &
           &
          AP$^{box}$ &
          AP$^{mask}$ &
          AP$^{box}$ &
          AP$^{mask}$ &
          AP$^{box}$ &
          AP$^{mask}$ &
          AP$^{box}$ &
          AP$^{mask}$ \\ \hhline{==========}
        FP &
          32/32 &
          46.0 &
          41.6 &
          48.5 &
          43.4 &
          50.4 &
          43.7 &
          51.9 &
          45.0 \\ \hline
        PTQ4ViT\cite{ptq4vit} &
          6/6 &
          5.8 &
          6.8 &
          6.5 &
          6.6 &
          14.7 &
          13.6 &
          12.5 &
          10.8 \\
        APQ-ViT\cite{apq_vit} &
          6/6 &
          45.4 &
          41.2 &
          47.9 &
          42.9 &
          48.6 &
          42.5 &
          50.5 &
          43.9 \\
        TSPTQ-ViT\cite{tsptq_vit} &
          6/6 &
          45.8 &
          \textbf{41.4} &
          \textbf{48.3} &
          \textbf{43.2} &
          \textbf{50.2} &
          43.5 &
          \textbf{51.8} &
          \textbf{44.8} \\
        \textbf{MPTQ-ViT} &
          6/6 &
          \textbf{45.9} &
          \textbf{41.4} &
          \textbf{48.3} &
          43.1 &
          \textbf{50.2} &
          \textbf{43.6} &
          \textbf{51.8} &
          \textbf{44.8} \\ \hline
        PTQ4ViT\cite{ptq4vit} &
          4/4 &
          6.9 &
          7.0 &
          26.7 &
          26.6 &
          14.7 &
          13.5 &
          0.5 &
          0.5 \\
        APQ-ViT\cite{apq_vit} &
          4/4 &
          23.7 &
          22.6 &
          44.7 &
          40.1 &
          27.2 &
          24.4 &
          47.7 &
          41.1 \\
        TSPTQ-ViT\cite{tsptq_vit} &
          4/4 &
          42.9 &
          39.3 &
          45.0 &
          40.7 &
          47.8 &
          41.6 &
          48.8 &
          42.5 \\
        \textbf{MPTQ-ViT} &
          4/4 &
          \textbf{44.2} &
          \textbf{40.2} &
          \textbf{47.3} &
          \textbf{42.7} &
          \textbf{49.2} &
          \textbf{42.7} &
          \textbf{50.8} &
          \textbf{44.2} \\ 
        \Xhline{1pt}
        \end{tabular}
        \vspace{-0.8em}
        \caption{Performance comparison of object detection on COCO dataset. AP$^{box}$ is the box average precision and AP$^{mask}$ is the mask average precision. Softmax and LayerNorm are preserved in FP. W/A denotes the bit-width for weight/activation.}
        \label{tab:object_detect}
        \vspace{-1.3em}
        \end{table*}
        \linespread{1.0}

    \subsection{Performance Comparison and Ablation Study} \label{sec:performance}
        
        \subsubsection {Single-Precision Quantization (SP)}\label{sec:sp}
        In \cref{tab:SP}, we present the top-1 accuracy under SP on the ImageNet dataset.
        For a fair comparison, Softmax and LayerNorm are preserved in FP.
        In 6-bit quantization, the proposed SQ-b+OPT-m achieves similar or even higher accuracy compared to prior works.
        For instance, our methods outperform TSPTQ-ViT \cite{tsptq_vit} by 1.52\% on DeiT-S.
        Though NoisyQuant \cite{noisyquant} achieves slightly higher performance than our works on Swin-S and Swin-B, it requires channel-wise quantization and addition computation during inference.
        Under 4-bit quantization, our methods show greater benefits.
        Specifically, we improve 23.35\% on DeiT-S and 11.74\% on ViT-B than TSPTQ-ViT \cite{tsptq_vit}.

        To validate the effectiveness of SQ-b and OPT-m, we conduct an ablation study, as shown in \cref{tab:ablation}.
        We implement SmoothQuant \cite{smoothquant} on the Baseline, denoted as SQ.
        From the results, SQ significantly improves the accuracy of the Baseline approach, indicating the outlier issue substantially impacts performance.
        After replacing SQ with SQ-b, we obtain a similar or even higher performance than SQ.
        Finally, by introducing the data-dependent quantizer, the models with SQ-b+OPT-m gain the best performance.

        To sum up, the proposed SQ-b and OPT-m exhibit accuracy improvements ranging from 0.90\% to 23.35\% compared with TSPTQ-ViT \cite{tsptq_vit} under 4-bit SP scenario. The results validate the effectiveness of enhancing data symmetry by SQ-b and deploying data-dependent SFs by OPT-m.

        \subsubsection {Mixed-Precision Quantization (MP)}\label{sec:mp}
        In \cref{tab:MP}, we demonstrate the top-1 accuracy under fully quantized MP on the ImageNet dataset. 
        The Baseline method shows that SP crashes under 5-bit quantization, emphasizing the necessity of MP for aggressive quantization.
        After introducing the proposed Greedy MP, there are substantial accuracy improvements across all types of models, particularly for the 5-bit scenario.
        While ViT-B using the Baseline method only receives 0.30\% accuracy under 5-bit quantization, the one with Greedy MP significantly enhances the accuracy to 79.66\%.
        Moreover, combining Greedy MP with SQ-b and OPT-m, denoted as MPTQ-ViT, leads to further accuracy improvements in most cases. 

        To evaluate the generalization of the proposed MPTQ-ViT, we extend it to object detection tasks on the COCO dataset, as shown in \cref{tab:object_detect}.
        From the results, we can observe PTQ4ViT \cite{ptq4vit} fails to maintain precision even under 6-bit quantization.
        Under the 4-bit scenario, APQ-ViT \cite{apq_vit} experiences significant performance degradation when Swin-T is used as the backbone. For instance, APQ-ViT \cite{apq_vit} obtains only 23.7 box AP and 22.6 mask AP on the Mask R-CNN framework with Swin-T.
        Lastly, both TSPTQ-ViT \cite{tsptq_vit} and MPTQ-ViT achieve near loss-less under 6-bit quantization, however, MPTQ-ViT surpasses TSPTQ-ViT across all models under 4-bit quantization.

        In a nutshell, Greedy MP enables aggressive quantization by exploiting layer-wise redundancy. 
        Compared with the Baseline approach on ImageNet, MPTQ-ViT exhibits accuracy improvements ranging from 0.76\% to 22.36\% under 6-bit MP and from 3.82\% to 78.14\% under 5-bit MP.
        

        \linespread{0.95}
        \begin{table}[t]
        \vspace{-0.3em}
            \centering
            \begin{tabular}{c|cccc}
            \Xhline{1pt}
            Method               & W/A   & ViT-L          & DeiT-B         & Swin-B         \\\hhline{=====}
            FP                   & 32/32 & 85.84          & 81.80          & 85.27          \\ \hline
            $\mu^+$ (O-Sup\cite{outlier_suppression+}) & 6/6   & 85.26          & \textbf{81.29} & 84.36          \\
            \textbf{$\mu^{Ours}$ (SQ-b)}        & 6/6   & \textbf{85.41} & 81.17          & \textbf{84.49} \\ \hline
            $\mu^+$ (O-Sup\cite{outlier_suppression+})  & 4/4   & 76.37          & 73.05          & \textbf{75.78} \\
            \textbf{$\mu^{Ours}$ (SQ-b)}        & 4/4   & \textbf{81.41} & \textbf{76.08} & 75.44 \\  
            \Xhline{1pt}
            \end{tabular}
            \vspace{-0.8em}
            \caption{Performance comparison between $\mu^+$ and $\mu^{Ours}$. Softmax and Layernorm are preserved in FP.}
            \label{tab:SQ-b_shift}
            \vspace{-1em}
        \end{table}
        \linespread{1.0}
        
        \linespread{0.95}
        \begin{table}[t]
            \centering
            \begin{tabular}{c|cccc}
            \Xhline{1pt}
            Method               & W/A   & ViT-L          & DeiT-B         & Swin-B         \\ \hhline{=====}
            FP                   & 32/32 & 85.84          & 81.80          & 85.27          \\ \hline
            $\epsilon^+$ (O-Sup\cite{outlier_suppression+}) & 6/6   & \textbf{85.58}          & 81.54 & 85.00          \\
            \textbf{$\epsilon^{Ours}$ (SQ-b)}        & 6/6   & 85.54 & \textbf{81.58}          & \textbf{85.13} \\ \hline
            $\epsilon^+$ (O-Sup\cite{outlier_suppression+}) & 4/4   & 3.28           & 60.25          & 77.55 \\
            \textbf{$\epsilon^{Ours}$ (SQ-b)}         & 4/4   & \textbf{7.97}  & \textbf{64.53} & \textbf{79.81}\\         
            \Xhline{1pt}
            \end{tabular}
            \vspace{-0.8em}
            \caption{Performance comparison between $\epsilon^+$ and $\epsilon^{Ours}$. Inputs of FC1 are quantized while other layers are preserved in FP.}
            \label{tab:SQ-b_smooth}
            \vspace{-1.8em}
        \end{table}
        \linespread{1.0}
        
    \subsection{Comparison with Value Redistribution for NLP} \label{sec:analysis_SQb}
        
        From here, we have confirmed the effectiveness of SQ-b in addressing the asymmetry. In this section, we further adapt Outlier Suppression+ \cite{outlier_suppression+} (denoted as O-Sup) to ViTs to compare with our approach. 
        Specifically, we modify $\mu_j$ and $\epsilon_j$ of SQ-b to $\mu_j^+=\frac{max(\bar X_j)+min(\bar X_j)}{2}$ and $\epsilon_j^+=\frac{max(\bar X_j-\mu_j)}{t_j}$, where $t_j$ is determined by grid search.
        We use $\mu^{Ours}$ and $\epsilon^{Ours}$ to denote $\mu$ and $\epsilon$ in SQ-b.
        
        We conduct the experiments under the SP scenario. 
        In the comparison of $\mu$, Softmax and LayerNorm are preserved in FP for accuracy preservation.
        In comparing $\epsilon$, we find $s_0$ by MinMax quantization rather than Hessian guided metric to align with \cite{outlier_suppression+}.
        We quantize the inputs of FC1 while preserving other layers in FP.    
        
\begin{figure}[t]
            \vspace{-0.7em}
            \centering
            \includegraphics[width=1\linewidth]{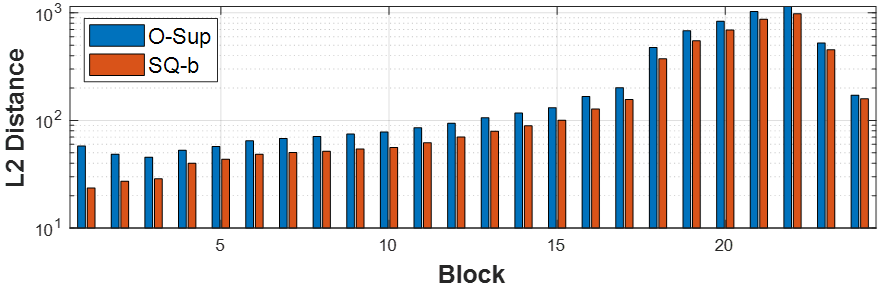}
            \vspace{-2.3em}
            \caption{L2 distance between $\mu$ and $\mu_r$ of ViT-L.}
           \label{fig:sumL2}
            \vspace{-2em}
        \end{figure}

        In \cref{tab:SQ-b_shift}, we present the comparison between $\mu^+$ and $\mu^{Ours}$.
        The results show that both approaches achieve similar accuracy under 6-bit quantization.
        However, when further quantized to 4-bit, SQ-b demonstrates greater robustness than \cite{outlier_suppression+}, improving accuracy by 5.04\% on ViT-L.
        To further analyze, we calculate the L2 distance between the $\mu$ calculated by the entire calibration data and the respective $\mu_r$ obtained from each individual calibration sample on ViT-L.
        Since a larger distance indicates higher variance in $\mu_r$, we prefer a smaller distance as it suggests $\mu$ better represents the general characteristics.
        From \cref{fig:sumL2}, we observe that SQ-b has a smaller L2 distance than O-Sup \cite{outlier_suppression+} across all blocks, indicating $\mu_r^{ours}$ owns lower variance and $\mu^{ours}$ is more representative.
        
        In \cref{tab:SQ-b_smooth}, we compare the effects of $\epsilon^+$ and $\epsilon^{Ours}$.
        Similar to the discussion of $\mu$, both methods preserve performance under 6-bit quantization and SQ-b outperforms O-Sup \cite{outlier_suppression+} under 4-bit case.
        For instance, SQ-b enhances the accuracy by 4.28\% compared to \cite{outlier_suppression+} on DeiT-B.
        We conjecture the accuracy drop of \cite{outlier_suppression+} is due to the over-fitting of $t_j$.
        Since image data is inherently more diverse than NLP data, a grid search based on a small calibration dataset may fail to well represent the varying data.

        The results demonstrate the parameters tailored for NLP tasks are not directly applicable to ViTs. Nevertheless, the proposed SQ-b method, designed specifically for ViTs, effectively improves model accuracy.
         
    \subsection{Visualization of Post-GeLU Quantization} \label{sec:analysis_OPTm}
        \begin{figure}[t]
            \centering
            \includegraphics[width=1\linewidth]{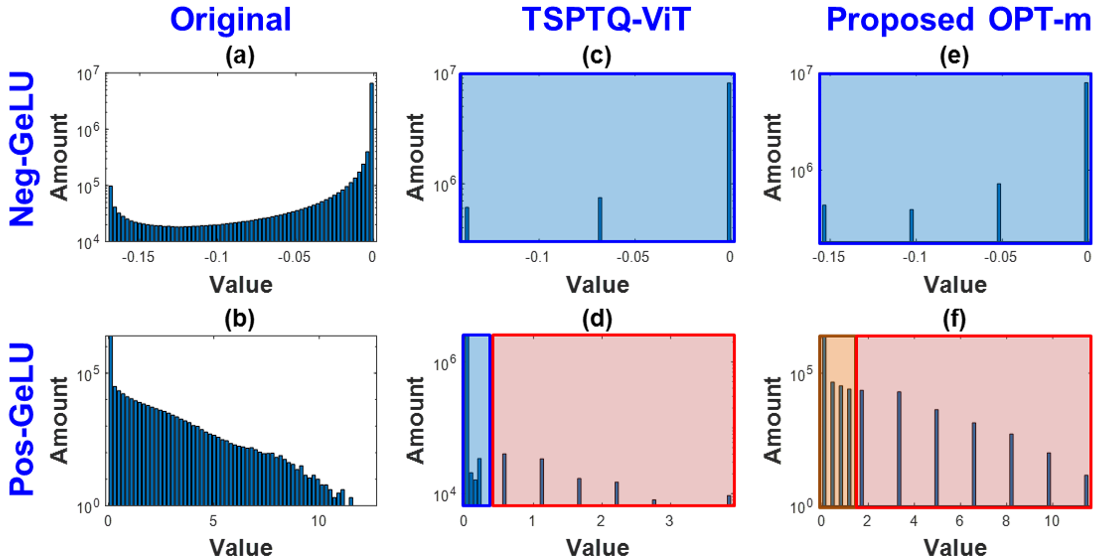}
            \vspace{-2.1em}
            \caption{Distribution of negative (Neg) and positive (Pos) post-GeLU values of $9^{th}$ blocks of DeiT-S under 4-bit quantization: (a)(b) original, (c)(d) TSPTQ-ViT \cite{tsptq_vit}, (e)(f) proposed OPT-m.}
           \label{fig:ptq_dist_v2sf_optm}
            \vspace{-1.1em}
        \end{figure}
        To compare TSPTQ-ViT \cite{tsptq_vit} with the proposed OPT-m, we visualize their value distribution under 4-bit quantization, as shown in \cref{fig:ptq_dist_v2sf_optm}.
        Values in each color region are represented by a specific SF. TSPTQ-ViT \cite{tsptq_vit} divides the values based on their magnitude while OPT-m divides them as in \cref{sec: OPT-m}.
        Comparing the negative post-GeLU values in \cref{fig:ptq_dist_v2sf_optm}(c) and (e), there are only 3 bins under TSPTQ-ViT \cite{tsptq_vit} with a SF of $0.0693$. However, the proposed OPT-m exhibits 4 bins with $s_0	\approx0.0515$, indicating higher precision.
        Next, we discuss the positive post-GeLU values in \cref{fig:ptq_dist_v2sf_optm}(d) and (f).
        While there are 4 bins in the small positive region under TSPTQ-ViT \cite{tsptq_vit}, the representative range is still significantly smaller compared to the entire positive region.
        Sharing the same SF for small magnitude values results in the negative region being represented by an unfitted SF and limits the small positive region to a narrow range.
        Moreover, the large SF in TSPTQ-ViT \cite{tsptq_vit} is fixed to be $8$ times the small SF, which is $0.5544$.
        Thus, the maximum value after quantization is $0.5544\cdot7=3.8808$, which is much lower than the original maximum value of $12.4909$.
        On the other hand, OPT-m uses a data-dependent strategy to determine the suitable SF ratio, which are 
        $m_0=3$ and $m_1=5$ in this case.
        Consequently, the SFs are $s_1=s_0\cdot2^3\approx0.4124$ for small positive values and $s_2=s_0\cdot2^5\approx1.6494$ for large positive values.
        The small positive region of OPT-m is broader than that of TSPTQ-ViT \cite{tsptq_vit}.
        Moreover, the maximum value after quantization is $1.6494\cdot7=11.5458$, significantly reducing the clamping loss.
        
        By dividing the values into multiple regions and using data-dependent SFs, OPT-m effectively enhances the precision of the post-quantization values.

    \subsection{Analysis of MPTQ-ViT Framework} 
    \label{sec:analysis_MP}
        \linespread{0.95}
        \begin{table}[t]
            \centering
            \begin{tabular}{c|cccc}
            \Xhline{1pt}
            Method               & W/A   & Swin-T          & Swin-S         & Swin-B         \\\hhline{=====}
            FP                   & 32/32 & 81.37          & 83.21          & 85.27          \\ \hline
            Baseline (SP) & 6/6   & 80.04          & 82.20 & 83.96          \\
            \textbf{Greedy MP}        & 6/6   & \textbf{80.49} & \textbf{82.61}          & \textbf{84.61} \\
            \Xhline{1pt}
            \end{tabular}
            \vspace{-0.8em}
            \caption{Top-1 accuracy on ImageNet dataset. Softmax is quantized as 8-bit.}
            \label{tab:MP_sm}
            \vspace{-1.9em}
        \end{table}
        \linespread{1.0}
        In this section, we analyze the results of Greedy MP to gain deeper insights. Specifically, we observe the quantized accuracy of Swin and then visualize the BW distribution.
        \begin{figure}[t]
            \centering
            \includegraphics[width=1\linewidth]{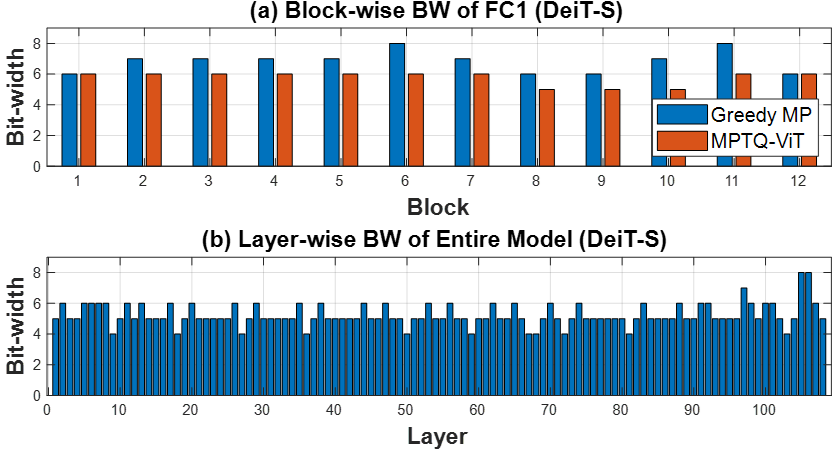}
            \vspace{-2.3em}
            \caption{(a) Block-wise BW distribution of FC1 in DeiT-S. (b) Layer-wise BW distribution of the entire DeiT-S model.}
           \label{fig:BW_dist}
        \vspace{-1.9em}
        \end{figure}
        
        In \cref{tab:MP}, though Swin-S and Swin-B achieve higher accuracy in FP than Swin-T, their accuracy is lower after quantization.
        We attribute this to the high sensitivity of Softmax to quantization and the compact architecture of Swin models.
        Additionally, the error accumulation has a greater impact on larger models, \ie Swin-S and Swin-B.
        To validate, we apply Greedy MP to layers other than Softmax while keeping the BW of Softmax fixed at 8 bits, as shown in \cref{tab:MP_sm}.
        From the results, the modified versions recover approximately 40\% accuracy, and Swin-S and Swin-B receive higher accuracy than Swin-T.
        This phenomenon highlights the challenges of quantizing Softmax and explains why many prior works ignore them.

        Next, we visualize the BW distribution under  5-bit MP.
        We compare the BW of inputs of FC1 in DeiT-S obtained from Greedy MP and MPTQ-ViT, as shown in \cref{fig:BW_dist}(a).
        From the result of Greedy MP, some FC1 layers are allocated relatively high BW, \eg 8-bit for the $6^{th}$ and $11^{th}$ blocks.
        Notably, after value redistribution by SQ-b, MPTQ-ViT perceives the difference and subsequently reduces the required BW of FC1.
        This result also indicates that SQ-b enables precision preservation with lower BW.
        Moreover, we demonstrate the layer-wise BW distribution of the entire DeiT model in \cref{fig:BW_dist}(b).
        Except the last few layers are allocated higher BW than the others, there is no obvious BW difference related to the layer's location.
        Instead, the variation in BW occurs within individual blocks, such as FC1 tends to have higher BW. 
        This suggests the operator types are more dominant in the BW allocation than the relative location in the model.
        
        In summary, our analysis of the BW distribution highlights the effectiveness of SQ-b and the influence of the operator types. These findings emphasize the ability of the proposed MPTQ-ViT to perceive the varying sensitivity to quantization and allocate BW to each layer accordingly.

\section{Conclusion}\label{sec:conclusion}
    This paper presents MPTQ-ViT, a novel mixed-precision post-training quantization scheme for ViTs. MPTQ-ViT incorporates SQ-b for symmetric distribution, OPT-m for data-dependent SFs, and Greedy MP for layer-wise BW allocation. Experimental results demonstrate the proposed methods achieve SOTA performance under SP and MP.

{\small
\bibliographystyle{unsrt}
\bibliography{egbib}

\begin{thebibliography}{10}

\bibitem{bert}
Jacob Devlin, Ming-Wei Chang, Kenton Lee, and Kristina Toutanova.
\newblock {BERT}: Pre-training of deep bidirectional transformers for language understanding.
\newblock In {\em NAACL}, 2019.

\bibitem{vit}
Alexey Dosovitskiy, Lucas Beyer, Alexander Kolesnikov, Dirk Weissenborn, Xiaohua Zhai, Thomas Unterthiner, Mostafa Dehghani, Matthias Minderer, Georg Heigold, Sylvain Gelly, Jakob Uszkoreit, and Neil Houlsby.
\newblock An image is worth 16x16 words: Transformers for image recognition at scale.
\newblock {\em ICLR}, 2021.

\bibitem{deit}
Hugo Touvron, Matthieu Cord, Matthijs Douze, Francisco Massa, Alexandre Sablayrolles, and Herv{\'e} J{\'e}gou.
\newblock Training data-efficient image transformers \& distillation through attention.
\newblock In {\em ICML}, 2021.

\bibitem{Swin}
Ze~Liu, Yutong Lin, Yue Cao, Han Hu, Yixuan Wei, Zheng Zhang, Stephen Lin, and Baining Guo.
\newblock Swin transformer: Hierarchical vision transformer using shifted windows.
\newblock In {\em ICCV}, 2021.

\bibitem{mask}
Kaiming He, Georgia Gkioxari, Piotr Doll{\'a}r, and Ross Girshick.
\newblock Mask r-cnn.
\newblock In {\em ICCV}, 2017.

\bibitem{cascade}
Zhaowei Cai and Nuno Vasconcelos.
\newblock Cascade r-cnn: Delving into high quality object detection.
\newblock In {\em CVPR}, 2018.

\bibitem{segformer}
Enze Xie, Wenhai Wang, Zhiding Yu, Anima Anandkumar, Jose~M Alvarez, and Ping Luo.
\newblock Segformer: Simple and efficient design for semantic segmentation with transformers.
\newblock In {\em NeurIPS}, 2021.

\bibitem{up_down}
Markus Nagel, Rana~Ali Amjad, Mart Van~Baalen, Christos Louizos, and Tijmen Blankevoort.
\newblock Up or down? adaptive rounding for post-training quantization.
\newblock In {\em ICML}, 2020.

\bibitem{opt_exponent}
Janghwan Lee and Jungwook Choi.
\newblock Optimizing exponent bias for sub-8bit floating-point inference of fine-tuned transformers.
\newblock In {\em AICAS}, 2022.

\bibitem{alps}
Hamed~F. Langroudi, Vedant Karia, Zachariah Carmichael, Abdullah Zyarah, Tej Pandit, John~L. Gustafson, and Dhireesha Kudithipudi.
\newblock Alps: Adaptive quantization of deep neural networks with generalized posits.
\newblock In {\em CVPRW}, 2021.

\bibitem{learnablMP}
Yu-Shan Tai, Cheng-Yang Chang, Chieh-Fang Teng, and An-Yeu~Andy Wu.
\newblock Learnable mixed-precision and dimension reduction co-design for low-storage activation.
\newblock In {\em SiPS}, 2022.

\bibitem{rethinking}
Zhaowei Cai and Nuno Vasconcelos.
\newblock Rethinking differentiable search for mixed-precision neural networks.
\newblock In {\em CVPR}, 2020.

\bibitem{GMPQ}
Ziwei Wang, Han Xiao, Jiwen Lu, and Jie Zhou.
\newblock Generalizable mixed-precision quantization via attribution rank preservation.
\newblock In {\em ICCV}, 2021.

\bibitem{evolutionary}
Zhenhua Liu, Xinfeng Zhang, Shanshe Wang, Siwei Ma, and Wen Gao.
\newblock Evolutionary quantization of neural networks with mixed-precision.
\newblock In {\em ICASSP}, 2021.

\bibitem{ptq_for_vit}
Zhenhua Liu, Yunhe Wang, Kai Han, Wei Zhang, Siwei Ma, and Wen Gao.
\newblock Post-training quantization for vision transformer.
\newblock In {\em NeurIPS}, 2021.

\bibitem{data_free_MP}
Vladimir Chikin and Mikhail Antiukh.
\newblock Data-free network compression via parametric non-uniform mixed precision quantization.
\newblock In {\em CVPR}, 2022.

\bibitem{apq}
Tianzhe Wang, Kuan Wang, Han Cai, Ji~Lin, Zhijian Liu, Hanrui Wang, Yujun Lin, and Song Han.
\newblock Apq: Joint search for network architecture, pruning and quantization policy.
\newblock In {\em CVPR}, 2020.

\bibitem{deep_com}
Song Han, Huizi Mao, and William~J Dally.
\newblock Deep compression: Compressing deep neural networks with pruning, trained quantization and huffman coding.
\newblock In {\em ICLR}, 2016.

\bibitem{xpruner}
Lu~Yu and Wei Xiang.
\newblock X-pruner: explainable pruning for vision transformers.
\newblock In {\em CVPR}, 2023.

\bibitem{joint_pruning}
Siyuan Wei, Tianzhu Ye, Shen Zhang, Yao Tang, and Jiajun Liang.
\newblock Joint token pruning and squeezing towards more aggressive compression of vision transformers.
\newblock In {\em CVPR}, 2023.

\bibitem{GreedyDR}
Yu-Shan Tai, Chieh-Fang Teng, Cheng-Yang Chang, and An-Yeu~Andy Wu.
\newblock Compression-aware projection with greedy dimension reduction for convolutional neural network activations.
\newblock In {\em ICASSP}, 2022.

\bibitem{feature_trans}
Brian Chmiel, Chaim Baskin, Evgenii Zheltonozhskii, Ron Banner, Yevgeny Yermolin, Alex Karbachevsky, Alex~M Bronstein, and Avi Mendelson.
\newblock Feature map transform coding for energy-efficient cnn inference.
\newblock In {\em IJCNN}, 2020.

\bibitem{joint_mp_dr}
Yu-Shan Tai, Cheng-Yang Chang, Chieh-Fang Teng, Yi-Ta Chen, and An-Yeu Wu.
\newblock Joint optimization of dimension reduction and mixed-precision quantization for activation compression of neural networks.
\newblock {\em IEEE Transactions on Computer-Aided Design of Integrated Circuits and Systems}, 2023.

\bibitem{qvit}
Zhexin Li, Tong Yang, Peisong Wang, and Jian Cheng.
\newblock Q-vit: Fully differentiable quantization for vision transformer.
\newblock In {\em NeurIPS}, 2022.

\bibitem{binaryvit}
Phuoc-Hoan~Charles Le and Xinlin Li.
\newblock Binaryvit: Pushing binary vision transformers towards convolutional models.
\newblock In {\em CVPR}, 2023.

\bibitem{tervit}
Sheng Xu, Yanjing Li, Teli Ma, Bohan Zeng, Baochang Zhang, Peng Gao, and Jinhu Lv.
\newblock Tervit: An efficient ternary vision transformer.
\newblock {\em arXiv preprint arXiv:2201.08050}, 2022.

\bibitem{Learnable}
Longguang Wang, Xiaoyu Dong, Yingqian Wang, Li~Liu, Wei An, and Yulan Guo.
\newblock Learnable lookup table for neural network quantization.
\newblock In {\em CVPR}, 2022.

\bibitem{bibert}
Haotong Qin, Yifu Ding, Mingyuan Zhang, Qinghua YAN, Aishan Liu, Qingqing Dang, Ziwei Liu, and Xianglong Liu.
\newblock Bi{BERT}: Accurate fully binarized {BERT}.
\newblock In {\em ICLR}, 2022.

\bibitem{qbert}
Sheng Shen, Zhen Dong, Jiayu Ye, Linjian Ma, Zhewei Yao, Amir Gholami, Michael~W Mahoney, and Kurt Keutzer.
\newblock Q-bert: Hessian based ultra low precision quantization of bert.
\newblock In {\em AAAI}, 2020.

\bibitem{ptq4vit}
Zhihang Yuan, Chenhao Xue, Yiqi Chen, Qiang Wu, and Guangyu Sun.
\newblock Ptq4vit: Post-training quantization for vision transformers with twin uniform quantization.
\newblock In {\em ECCV}, 2022.

\bibitem{pd_quant}
Jiawei Liu, Lin Niu, Zhihang Yuan, Dawei Yang, Xinggang Wang, and Wenyu Liu.
\newblock Pd-quant: Post-training quantization based on prediction difference metric.
\newblock In {\em CVPR}, 2023.

\bibitem{tsptq_vit}
Yu-Shan Tai, Ming-Guang Lin, and An-Yeu~Andy Wu.
\newblock Tsptq-vit: Two-scaled post-training quantization for vision transformer.
\newblock In {\em ICASSP}, 2023.

\bibitem{smoothquant}
Guangxuan Xiao, Ji~Lin, Mickael Seznec, Hao Wu, Julien Demouth, and Song Han.
\newblock Smoothquant: Accurate and efficient post-training quantization for large language models.
\newblock In {\em ICML}, 2023.

\bibitem{outlier_suppression+}
Xiuying Wei, Yunchen Zhang, Yuhang Li, Xiangguo Zhang, Ruihao Gong, Jinyang Guo, and Xianglong Liu.
\newblock Outlier suppression+: Accurate quantization of large language models by equivalent and optimal shifting and scaling.
\newblock {\em arXiv preprint arXiv:2304.09145}, 2023.

\bibitem{apq_vit}
Yifu Ding, Haotong Qin, Qinghua Yan, Zhenhua Chai, Junjie Liu, Xiaolin Wei, and Xianglong Liu.
\newblock Towards accurate post-training quantization for vision transformer.
\newblock In {\em ACMMM}, 2022.

\bibitem{fqvit}
Yang Lin, Tianyu Zhang, Peiqin Sun, Zheng Li, and Shuchang Zhou.
\newblock Fq-vit: Post-training quantization for fully quantized vision transformer.
\newblock In {\em IJCAI}, 2022.

\bibitem{attention}
Ashish Vaswani, Noam Shazeer, Niki Parmar, Jakob Uszkoreit, Llion Jones, Aidan~N Gomez, {\L}ukasz Kaiser, and Illia Polosukhin.
\newblock Attention is all you need.
\newblock In {\em NeurIPS}, 2017.

\bibitem{noisyquant}
Yijiang Liu, Huanrui Yang, Zhen Dong, Kurt Keutzer, Li~Du, and Shanghang Zhang.
\newblock Noisyquant: Noisy bias-enhanced post-training activation quantization for vision transformers.
\newblock In {\em CVPR}, 2023.

\bibitem{imagenet}
Olga Russakovsky, Jia Deng, Hao Su, Jonathan Krause, Sanjeev Satheesh, Sean Ma, Zhiheng Huang, Andrej Karpathy, Aditya Khosla, Michael Bernstein, Alexander~C. Berg, and Li~Fei-Fei.
\newblock {ImageNet Large Scale Visual Recognition Challenge}.
\newblock {\em IJCV}, 115:211--252, 2015.

\bibitem{timm}
Ross Wightman.
\newblock Pytorch image models.
\newblock \url{https://github.com/rwightman/pytorch-image-models}, 2019.

\bibitem{coco}
Tsung-Yi Lin, Michael Maire, Serge Belongie, James Hays, Pietro Perona, Deva Ramanan, Piotr Doll{\'a}r, and C~Lawrence Zitnick.
\newblock Microsoft coco: Common objects in context.
\newblock In {\em ECCV}, 2014.

\end{thebibliography}
}

\clearpage
\setcounter{section}{0}
\setcounter{table}{0}
\setcounter{figure}{0}
\setcounter{equation}{0}
\renewcommand\thesubsection{\Alph{subsection}}
\section*{Supplementary Material}
\subsection{Visualization of Post-GeLU Activations}\label{sec:post_gelu}
    \begin{figure}[ht]
        \centering
        \includegraphics[width=1\linewidth]{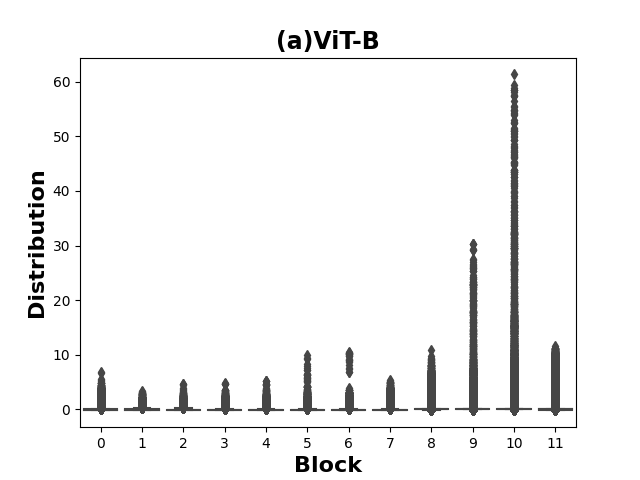}
        \includegraphics[width=1\linewidth]{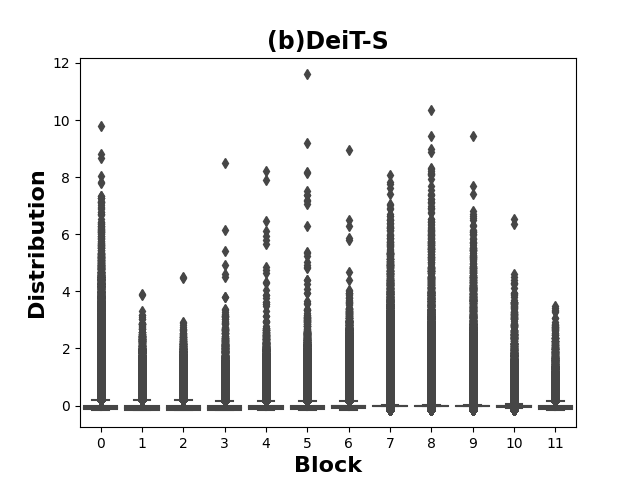}
        \caption{Box plots of block-wise post-GeLU values on (a) ViT-B and (b) DeiT-S.}
       \label{fig:box}
    \end{figure}
    In \cref{fig:box}, we present the box plots of block-wise post-GeLU values on ViT-B and DeiT-S. 
    The figures demonstrate highly variant distribution across different blocks, with the difference between the maximums 
     of the $1^{st}$ and $10^{th}$ blocks in ViT-B reaching up to $10\times$.
    Additionally, the distribution varies with the model, as demonstrated by the distinct box plots of ViT-B and DeiT-S.
    The diverse distribution across blocks and models highlights the importance of a data-dependent quantizer, as hand-crafted designs may fail to adapt to such varied conditions. 
    The limited generalizability of \cite{ptq4vit, tsptq_vit} also explains their performance discrepancies on different models, as shown in \cref{tab:SP}.
    In contrast, the proposed OPT-m can adapt to various data distributions and effectively preserve performance across different model types.
\subsection{Ablation Study of Greedy MP Metric}\label{sec:ablation_MP}
    \begin{table}[ht]
            \centering
            \begin{tabular}{c|cccc}
            \Xhline{1pt}
            Method          & W/A   & ViT-L          & DeiT-B         & Swin-B         \\ \hhline{=====}
            FP              & 32/32 & 85.84          & 81.80          & 85.27          \\ \hline
            Baseline (SP)        & 5/5   & 80.02          & 58.53          & 8.74          \\
            SQNR only    & 5/5   & 83.46          & 75.50          & 9.82          \\
            \textbf{Greedy MP} & 5/5   & \textbf{83.61} & \textbf{76.44} & \textbf{41.69}\\ \Xhline{1pt}
            \end{tabular}
            \caption{Ablation study of the selection metric of Greedy MP on ImageNet dataset.}
            \label{tab:ql_ablation}
        \end{table}
    In this section, we evaluate the effectiveness of the selection metric $\alpha_l$ mentioned in \cref{sec: OPT-m}.
    The proposed metric considers both model performance and compressibility, achieving a balanced trade-off between the two factors. 
    To validate its effectiveness, we conduct an ablation study by using SQNR only as the selection metric:
    \begin{equation}
            \alpha_l = SQNR_{b-1}(X_l).
            \label{eq:def_metric_sqnr_only}
        \end{equation}
    The comparison results in \cref{tab:ql_ablation} show that using $\alpha_l$ with SQNR only results in lower accuracy compared to the one further considering compressibility (Greedy MP). 
    This indicates the importance of considering both performance and compressibility in the selection metric.
    Note that using a selection metric solely based on the number of elements in each layer is unsuitable. 
    Due to the constant characteristic of the number of elements, the quantization process would repeatedly target the same layer.
    By incorporating both factors in our Greedy MP approach, we achieve a more effective and balanced layer-wise bit-width allocation for quantization.


\end{document}